# Continuous-Time Visual-Inertial Odometry for Event Cameras

Elias Mueggler, Guillermo Gallego, Henri Rebecq, and Davide Scaramuzza

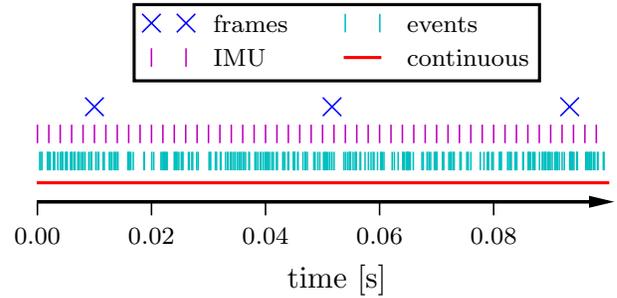

Fig. 1: While the frames and inertial measurements arrive at a constant rate, events are transmitted asynchronously and at much higher frequency. We model the trajectory of the combined camera-IMU sensor as continuous in time, which allows direct integration of all measurements using their precise timestamps.

*Abstract*—Event cameras are bio-inspired vision sensors that output pixel-level brightness changes instead of standard intensity frames. They offer significant advantages over standard cameras, namely a very high dynamic range, no motion blur, and a latency in the order of microseconds. However, due to the fundamentally different structure of the sensor's output, new algorithms that exploit the high temporal resolution and the asynchronous nature of the sensor are required. Recent work has shown that a continuous-time representation of the event camera pose can deal with the high temporal resolution and asynchronous nature of this sensor in a principled way. In this paper, we leverage such a continuous-time representation to perform visual-inertial odometry with an event camera. This representation allows direct integration of the asynchronous events with micro-second accuracy and the inertial measurements at high frequency. The event camera trajectory is approximated by a smooth curve in the space of rigid-body motions using cubic splines. This formulation significantly reduces the number of variables in trajectory estimation problems. We evaluate our method on real data from several scenes and compare the results against ground truth from a motion-capture system. We show that our method provides improved accuracy over the result of a state-of-the-art visual odometry method for event cameras. We also show that both the map orientation and scale can be recovered accurately by fusing events and inertial data. To the best of our knowledge, this is the first work on visual-inertial fusion with event cameras using a continuous-time framework.

## I. Introduction

EVENT cameras, such as the Dynamic Vision Sensor (DVS) [1], the DAVIS [2] or the ATIS [3], work very differently from a traditional camera. They have *independent* pixels that only send information (called "events") in presence of brightness changes in the scene at the time they occur. Thus, the output is not an intensity image but a stream of asynchronous events at *micro*-second resolution, where each event consists of its space-time coordinates and the *sign* of the brightness change (i.e., no intensity). Event cameras have numerous advantages over standard cameras: a latency in the order of microseconds, low power consumption, and a very high dynamic range ($130\,\text{dB}$ compared to $60\,\text{dB}$ of standard cameras). Most importantly, since all the pixels are independent, such sensors do not suffer from motion blur. However, because the output it produces—an event stream—is fundamentally different from video streams of standard cameras, new algorithms are required to deal with these data.

The authors are with the Robotics and Perception Group, Dept. of Informatics, University of Zurich, and Dept. of Neuroinformatics, University of Zurich and ETH Zurich, Switzerland—http://rpg.ifi.uzh.ch. This work was supported by the Swiss National Center of Competence in Research (NCCR) Robotics, through the Swiss National Science Foundation, by the SNSF-ERC Starting Grant, the DARPA FLA Program, the Qualcomm Innovation Fellowship and the UZH Forschungskredit.

In this paper, we aim to use an event camera in combination with an Inertial Measurement Unit (IMU) for ego-motion estimation. This task, called Visual-Inertial Odometry (VIO), has important applications in various fields, such as mobile robotics and augmented/virtual reality (AR/VR) applications.

The approach provided by traditional visual odometry frameworks, which estimate the camera pose at discrete times (naturally, the times at which images are acquired), is no longer appropriate for event cameras, mainly due to two issues. First, a single event does not contain enough information to estimate the six degrees of freedom (DOF) pose of a calibrated camera. Second, it is not appropriate to simply consider several events for determining the pose using standard computer-vision techniques, such as PnP [4], because the events typically all have different timestamps, and so the resulting pose will not correspond to any particular time (see Fig. 1). Third, an event camera can easily transmit up to several million events per second, and, therefore, it can become intractable to estimate the pose of the event camera at the discrete times of all events due to the rapidly growing size of the state vector needed to represent all such poses.

To tackle the above-mentioned issues, we adopt a continuous-time framework [5]. Regarding the first two issues, an explicit continuous temporal model is a natural representation of the pose trajectory $\mathtt{T}(t)$ of the event camera since it unambiguously relates each event, occurring at time $t_k$, with its corresponding pose, $\mathtt{T}(t_k)$. To solve the third issue, the trajectory is described by a smooth parametric model, with significantly fewer parameters than events, hence achieving state space size reduction and computational efficiency. For example, to remove unnecessary states for the estimation of the trajectory of dynamic objects, [6] proposed to use cubic splines, reporting state-space size compression of 70–90 %. Cubic splines [7] or, in more general, Wavelets [8] are common basis functions for continuous-time trajectories. The continuous-time framework was also motivated to allow data fusion of multiple sensors working at different rates and to enable increased temporal resolution [6]. This framework has been applied to camera-IMU fusion [7], [5], rolling-shutter cameras [5], actuated lidar [9], and RGB-D rolling-shutter cameras [10].

*Contribution*

The use of a continuous-time framework for ego motion estimation with event cameras was first introduced in our previous conference paper [11]. In the present paper, we extend [11] in several ways:

- While in [11] we used the continuous-time framework for trajectory estimation of an event camera, here we tackle the problem of trajectory estimation of a combined event-camera and IMU sensor. We show that the assimilation of inertial data allows us ($i$) to produce more accurate trajectories than with visual data alone and ($ii$) to estimate the absolute scale and orientation (alignment with respect to gravity).
- While [11] was limited to line-based maps, we extend the approach to work on natural scenes using point-based maps.
- We also show that our approach can be used to refine the poses estimated by an event-based visual odometry method [12].
- We demonstrate the capabilities of the extended approach with new experiments, including natural scenes.

This paper focuses on ($i$) presenting a direct way in which raw event and IMU measurements can be fused for VIO using a continuous-time framework and ($ii$) showing the accuracy of the approach. In particular, we demonstrate the accuracy on post-processing, full-smoothing camera trajectory (taking into account the full history of measurements).

The paper is organized as follows: Section II briefly introduces the principle of operation of event cameras, Section III reviews previous work on ego-motion estimation with event cameras, Sections IV to VI present our method for continuous-time trajectory optimization using visual-inertial event data fusion, Section VII presents the experiments carried out using the event camera to track two types of maps (point-based and line-based), Section VIII discusses the results, and Section IX draws final conclusions.

## II. EVENT CAMERAS

Standard cameras acquire frames (i.e., images) at fixed rates. On the other hand, event cameras such as the DAVIS [2] (Fig. 2) have independent pixels that output brightness changes (called "events") asynchronously, at the time they occur. Specifically, if $L(\mathbf{u}, t) \doteq \log I(\mathbf{u}, t)$ is the logarithmic brightness or intensity at pixel $\mathbf{u} = (x, y)^\top$ in the image plane, the DAVIS generates an event $e_k \doteq \langle x_k, y_k, t_k, p_k \rangle$ if the change in logarithmic brightness at pixel $\mathbf{u}_k = (x_k, y_k)^\top$ reaches a threshold $C$ (typically 10-15% relative brightness change):

$$\Delta L(\mathbf{u}_k, t_k) \doteq L(\mathbf{u}_k, t_k) - L(\mathbf{u}_k, t_k - \Delta t) = p_k C, \quad (1)$$

where $t_k$ is the timestamp of the event, $\Delta t$ is the time since the previous event at the same pixel $\mathbf{u}_k$ and $p_k \in \{-1, +1\}$ is the polarity of the event (the sign of the brightness change). Events are timestamped and transmitted asynchronously at the time they occur using a sophisticated digital circuitry.

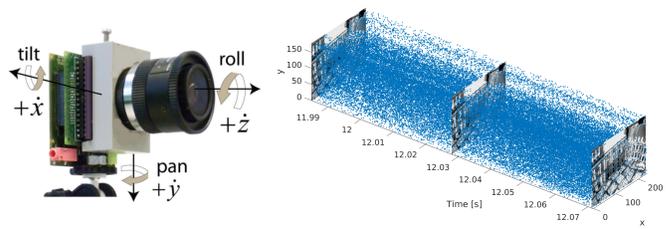

Fig. 2: The DAVIS camera from iniLabs (Figure adapted from [13]) and visualization of its output in space-time. Blue dots mark individual asynchronous events. The polarity of the events is not shown.

Event cameras have the same optics as traditional perspective cameras, therefore, standard camera models (e.g., pinhole) still apply. In this work, we use the DAVIS240C [2] that provides events and global-shutter images from the same physical pixels. In addition to the events and images, it contains a synchronized IMU. This work solely uses the images for camera calibration, initialization and visualization purposes. The sensor's spatial resolution is $240 \times 180$ pixels and it is connected via USB. A visualization of the output of the DAVIS is shown in Fig. 2. An additional advantage of the DAVIS is its very high dynamic range of 130 dB (compared to 60 dB of high quality traditional image sensors).

## III. RELATED WORK: EGO-MOTION ESTIMATION WITH EVENT CAMERAS

A particle-filter approach for robot self-localization using the DVS was introduced in [14] and later extended to SLAM in [15]. However, the system was limited to planar motions and planar scenes parallel to the plane of motion, and the scenes consisted of B&W line patterns.

In several works, conventional vision sensors have been attached to the event camera to simplify the ego-motion estimation problem. For example, [16] proposed an event-based probabilistic framework to update the relative pose of a DVS with respect to the last frame of an attached standard camera. The 3D SLAM system in [17] relied on a frame-based RGB-D camera attached to the DVS to provide depth estimation, and thus build a voxel grid map that was used for pose tracking. The system in [18] used the intensity images from the DAVIS camera to detect features that were tracked using the events and fed into a 3D visual odometry pipeline.

Robot localization in 6-DOF with respect to a map of B&W lines was demonstrated using a DVS, without additional sensing, during high-speed maneuvers of a quadrotor [19], where rotational speeds of up to $1{,}200\,°/s$ were measured. In natural scenes, [20] presented a probabilistic filter to track high-speed 6-DOF motions with respect to a map containing both depth and brightness information.

A system with two probabilistic filters operating in parallel was presented in [21] to estimate the rotational motion of a DVS and reconstruct the scene brightness in a high-resolution panorama. The system was extended in [22] using three filters that operated in parallel to estimate the 6-DOF pose of the event camera, and the depth and brightness of the scene. Recently, [12] presented a geometric parallel-tracking-and-mapping approach for 6-DOF pose estimation and 3D reconstruction with an event camera in natural scenes.

All previous methods operate in an event-by-event basis or in a groups-of-events basis [23], [24], [25], producing estimates of the event camera pose in a discrete manner. More recently, methods have been proposed that combine an event camera with an IMU rigidly attached for 6-DOF motion estimation [26], [27], [28]. These methods operate by processing small groups of events from which features (such as Harris [29] or FAST [30], [31]) can be tracked and fed into standard geometric VIO algorithms (MSCKF [32], OKVIS [33], optimization with pre-integrated IMU factors [34]). Yet again, these methods output a discrete set of camera-IMU poses.

This paper takes a different approach from previous methods. Instead of producing camera poses at discrete times, we estimate the trajectory of the camera as a continuous-time entity (represented by a set of control poses and interpolated using continuous-time basis functions [5]). The continuous-time representation allows fusing event and inertial data in a principled way, taking into account the precise timestamps of the measurements, without any approximation. Moreover, the representation is compact, using few parameters (control poses) to assimilate thousands of measurements. Finally, from a more technical point of view, classical VIO algorithms have separate estimates in the state vector for the pose and the linear velocity, which is prone to inconsistencies. In contrast, the continuous-time framework offers the advantage that both pose and linear velocity are derived from a unique, consistent trajectory representation.

## IV. CONTINUOUS-TIME SENSOR TRAJECTORIES

Traditional visual odometry and SLAM formulations use a discrete-time approach, i.e., the camera pose is calculated at the time the image was acquired. Recent works have shown that, for high-frequency data, a continuous-time formulation is preferable to keep the size of the optimization problem bounded [7], [5]. Temporal basis functions, such as B-splines, were proposed for camera-IMU calibration, where the frequencies of the two sensor modalities differ by an order of magnitude. While previous approaches use continuous-time representations mainly to reduce the computational complexity, in the case of an event-based sensor this representation is required to cope with the asynchronous nature of the events. Unlike a standard camera image, an event does not carry enough information to estimate the sensor pose by itself. A continuous-time trajectory can be evaluated at any time, in particular at each event's and inertial measurement's timestamp, yielding a well-defined pose and derivatives for every event. Thus, our method is not only computationally effective, but it is also necessary for a proper formulation.

### A. Camera Pose Transformations

Following [5], we represent camera poses by means of rigid-body motions using Lie group theory. A pose $\mathtt{T} \in SE(3)$, represented by a $4 \times 4$ transformation matrix

$$\mathtt{T} \doteq \begin{bmatrix} \mathtt{R} & \mathbf{t} \\ \mathbf{0}^\top & 1 \end{bmatrix} \quad (2)$$

(with rotational and translational components $\mathtt{R} \in SO(3)$ and $\mathbf{t} \in \mathbb{R}^3$, respectively) can be parametrized using exponential coordinates according to

$$\mathtt{T} = \exp(\widehat{\boldsymbol{\xi}}), \quad \text{for some } \textit{twist} \quad \widehat{\boldsymbol{\xi}} \doteq \begin{bmatrix} \widehat{\boldsymbol{\beta}} & \boldsymbol{\alpha} \\ \mathbf{0}^\top & 0 \end{bmatrix} \quad (3)$$

(with $\boldsymbol{\alpha} \in \mathbb{R}^3$, and $\widehat{\boldsymbol{\beta}}$ the $3 \times 3$ skew-symmetric matrix associated to vector $\boldsymbol{\beta} \in \mathbb{R}^3$, such that $\widehat{\boldsymbol{\beta}} \mathbf{a} = \boldsymbol{\beta} \times \mathbf{a}, \forall \boldsymbol{\beta}, \mathbf{a} \in \mathbb{R}^3$).

Every rigid-body motion $\mathtt{T} \in SE(3)$ can be written as (3), but the resulting twist $\widehat{\boldsymbol{\xi}}$ may not be unique [35, p. 33]. To avoid this ambiguity, specially for representing a continuously-varying pose $\mathtt{T}(t)$ (i.e., a camera trajectory), we adopt a local-charts approach on $SE(3)$, which essentially states that we represent short segments of the camera trajectory $\mathtt{T}(t)$ by means of an anchor pose $\mathtt{T}_a$ and an incremental motion with respect to it: $\mathtt{T}(t) = \mathtt{T}_a \exp(\widehat{\boldsymbol{\xi}}(t))$, with small matrix norm $\|\widehat{\boldsymbol{\xi}}(t)\|$. In addition, this approach is free from parametrization singularities. Closed-form formulas for the exponential map (3) and its inverse, the log map, are given in [35].

### B. Cubic Spline Camera Trajectories in SE(3)

We use B-splines to represent continuous-time trajectories in $SE(3)$ for several reasons: they $(i)$ are smooth ($C^2$ continuity in case of cubic splines), $(ii)$ have local support, $(iii)$ have analytical derivatives and integrals, $(iv)$ interpolate the pose at any point in time, thus enabling data fusion from both asynchronous and synchronous sensors with different rates.

The continuous trajectory of the event camera is parametrized by control poses $\mathtt{T}_{w,i}$ at times $\{t_i\}_{i=0}^n$, where $\mathtt{T}_{w,i}$ is the transformation from the event-camera coordinate system at time $t_i$ to a world coordinate system ($w$). Due to the locality of the cubic B-spline basis, the value of the spline curve at any time $t$ only depends on four control poses. For $t \in [t_i, t_{i+1})$ such control poses occur at times $\{t_{i-1}, \ldots, t_{i+2}\}$. Following the cumulative cubic B-splines formulation [5], which is illustrated in Fig. 3, we use one absolute pose, $\mathtt{T}_{w,i-1}$, and three incremental poses, parameterized by twists (3) $\widehat{\boldsymbol{\xi}}_q \equiv \Omega_q$. More specifically, the spline trajectory is given by

$$\mathtt{T}_{w,s}(u(t)) \doteq \mathtt{T}_{w,i-1} \prod_{j=1}^{3} \exp\left(\tilde{\mathbf{B}}_j(u(t))\, \Omega_{i+j-1}\right), \quad (4)$$

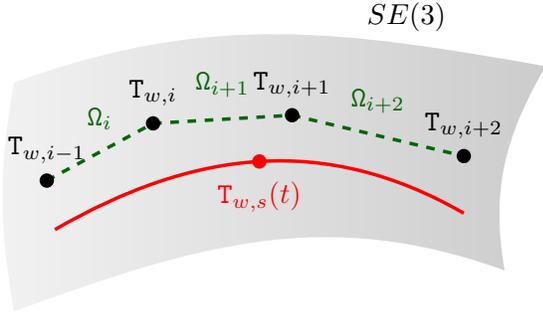

Fig. 3: Geometric interpretation of the cubic spline interpolation given by formula (4). The cumulative formulation uses one absolute control pose $\mathtt{T}_{w,i-1}$ and three incremental control poses $\Omega_i, \Omega_{i+1}, \Omega_{i+2}$ to compute the interpolated pose $\mathtt{T}_{w,s}$.

where we assume that the control poses are uniformly spaced in time [5], at $t_i = i\Delta t$. Thus $u(t) = (t - t_i)/\Delta t \in [0, 1)$ is used in the cumulative basis functions for the B-splines,

$$\tilde{\mathbf{B}}(u) = \mathbf{C} \begin{bmatrix} 1 \\ u \\ u^2 \\ u^3 \end{bmatrix}, \quad \mathbf{C} = \frac{1}{6} \begin{bmatrix} 6 & 0 & 0 & 0 \\ 5 & 3 & -3 & 1 \\ 1 & 3 & 3 & -2 \\ 0 & 0 & 0 & 1 \end{bmatrix}. \quad (5)$$

In (4), $\tilde{\mathbf{B}}_j$ is the $j$-th entry (0-based) of vector $\tilde{\mathbf{B}}$. The incremental pose from time $t_{i-1}$ to $t_i$ is encoded by the twist

$$\Omega_i = \log(\mathtt{T}_{w,i-1}^{-1} \mathtt{T}_{w,i}). \quad (6)$$

Temporal derivatives of the spline trajectory (4) are given in Appendix A. Analytical derivatives of $\mathtt{T}_{w,s}(u)$ with respect to the control poses are provided in the supplementary material of [10], which are based on [36].

### C. Generative Model for Visual and Inertial Observations

A continuous trajectory model allows us to compute the velocity and acceleration of the event camera at any time. These quantities can be compared against IMU measurements and the resulting mismatch can be used to refine the modeled trajectory. The predictions of the IMU measurements (angular velocity $\boldsymbol{\omega}$ and linear acceleration $\mathbf{a}$) are given by [5]

$$\hat{\boldsymbol{\omega}}(u) \doteq \left(\mathtt{R}_{w,s}^\top(u) \dot{\mathtt{R}}_{w,s}(u)\right)^\vee + \mathbf{b}_\omega, \quad (7)$$
$$\hat{\mathbf{a}}(u) \doteq \mathtt{R}_{w,s}^\top(u)\left(\ddot{\mathbf{s}}_w(u) + \mathbf{g}_w\right) + \mathbf{b}_a, \quad (8)$$

where $\dot{\mathtt{R}}_{w,s}(u)$ is the upper-left $3 \times 3$ sub-matrix of $\dot{\mathtt{T}}_{w,s}(u)$, and $\ddot{\mathbf{s}}_w(u)$ is the upper-right $3 \times 1$ sub-matrix of $\ddot{\mathtt{T}}_{w,s}(u)$. $\mathbf{b}_\omega$ and $\mathbf{b}_a$ are the gyroscope and accelerometer biases, and $\mathbf{g}_w$ is the acceleration due to gravity in the world coordinate system. The vee operator $[\cdot]^\vee$, inverse of the lift operator $\hat{\cdot}$, maps a $3\times 3$ skew-symmetric matrix to its corresponding vector [35, p. 18].

## V. SCENE MAP REPRESENTATION

To focus on the event-camera trajectory estimation problem, we assume that the map of the scene is given. Specifically, the 3D map $\mathcal{M}$ is either a set of points or line segments. We provide experiments using both geometric primitives.

In case of a map consisting of a set of points

$$\mathcal{M} = \{\mathbf{X}_i\}, \quad (9)$$

since events are caused by the apparent motion of edges, each 3D point $\mathbf{X}_i$ represents a scene edge. Given a $3 \times 4$ projection matrix $\mathtt{P}$ modeling the perspective projection carried out by the event camera, the event coordinates are, in homogeneous coordinates, $\mathbf{u}_i \propto \mathtt{P} \mathbf{X}_i$.

In the case of lines, the map is

$$\mathcal{M} = \{\ell_j\}, \quad (10)$$

where each line segment $\ell_j$ is parametrized by its start and end points $\mathbf{X}_j^s, \mathbf{X}_j^e \in \mathbb{R}^3$. The lines of the map $\mathcal{M}$ can be projected to the image plane by projecting the endpoints of the segments. The homogeneous coordinates of the projected line through the $j$-th segment are $l_j \propto (\mathtt{P} \mathbf{X}_j^s) \times (\mathtt{P} \mathbf{X}_j^e)$.

## VI. CAMERA TRAJECTORY OPTIMIZATION

In this section, we formulate the camera trajectory estimation problem from visual and inertial data in a probabilistic framework and derive the maximum likelihood criterion (Section VI-A). To find a tractable solution, we reduce the problem dimensionality (Section VI-B) by using the parametrized cubic spline trajectory representation introduced in Section IV-B.

### A. Probabilistic Approach

In general, the trajectory estimation problem over an interval $[0, T]$ can be cast in a probabilistic form [7], seeking an estimate of the joint posterior density $p(\mathbf{x}(t)|\mathcal{M}, \mathcal{Z})$ of the state $\mathbf{x}(t)$ (event camera trajectory) over the interval, given the map $\mathcal{M}$ and all visual-inertial measurements, $\mathcal{Z} = \mathcal{E} \cup \mathcal{W} \cup \mathcal{A}$, which consists of: events $\mathcal{E} \doteq \{\mathbf{e}_k\}_{k=1}^N$ (where $\mathbf{e}_k = (x_k, y_k)^\top$ is the event location at time $t_k$), angular velocities $\mathcal{W} \doteq \{\boldsymbol{\omega}_j\}_{j=1}^M$ and linear accelerations $\mathcal{A} \doteq \{\mathbf{a}_j\}_{j=1}^M$. Using Bayes' rule, and assuming that the map is independent of the event camera trajectory, we may rewrite the posterior as

$$p(\mathbf{x}(t) \mid \mathcal{M}, \mathcal{E}, \mathcal{W}, \mathcal{A}) \propto p(\mathbf{x}(t)) \, p(\mathcal{E}, \mathcal{W}, \mathcal{A} \mid \mathbf{x}(t), \mathcal{M}). \quad (11)$$

In the absence of prior belief for the state, $p(\mathbf{x}(t))$, the optimal trajectory is the one maximizing the likelihood $p(\mathcal{E}, \mathcal{W}, \mathcal{A} \mid \mathbf{x}(t), \mathcal{M})$. Assuming that the measurements $\mathcal{E}, \mathcal{W}, \mathcal{A}$ are independent of each other given the trajectory and the map, and using the fact that the inertial measurements do not depend on the map, the likelihood factorizes:

$$p(\mathcal{E}, \mathcal{W}, \mathcal{A}|\mathbf{x}(t), \mathcal{M}) = p(\mathcal{E}|\mathbf{x}(t), \mathcal{M}) \, p(\mathcal{W}|\mathbf{x}(t)) \, p(\mathcal{A} \mid \mathbf{x}(t)). \quad (12)$$

The first term in (12) comprises the visual measurements only. Under the assumption that the measurements $\mathbf{e}_k$ are independent of each other (given the trajectory and the map) and that the measurement error in the image coordinates of the events follows a zero-mean Gaussian distribution with variance $\sigma_e^2$, we have

$$\log\big(p(\mathcal{E} \mid \mathbf{x}(t), \mathcal{M})\big) \tag{13}$$

$$= \log\left(\prod_k p(\mathbf{e}_k|\mathbf{x}(t_k), \mathcal{M})\right) \tag{14}$$

$$= \log\left(\prod_k K_1 \exp\left(-\frac{\|\mathbf{e}_k - \hat{\mathbf{e}}_k(\mathbf{x}(t_k), \mathcal{M})\|^2}{2\sigma_e^2}\right)\right) \tag{15}$$

$$= \tilde{K}_1 - \frac{1}{2}\sum_k \frac{1}{\sigma_e^2}\|\mathbf{e}_k - \hat{\mathbf{e}}_k(\mathbf{x}(t_k), \mathcal{M})\|^2 \tag{16}$$

where $K_1 \doteq 1/\sqrt{2\pi\sigma_e^2}$ and $\tilde{K}_1 \doteq \sum_k \log K_1$ are constants (i.e., independent of the state $\mathbf{x}(t)$). Let us denote by $\hat{\mathbf{e}}_k(\mathbf{x}(t_k), \mathcal{M})$ the predicted value of the event location computed using the state $\mathbf{x}(t)$ and the map $\mathcal{M}$. Such a prediction is a point on one of the projected 3D primitives: in case of a map of points (9), $\hat{\mathbf{e}}$ is the projected point, and the norm in (16) is the standard reprojection error between two points; in case of a map of 3D line segments (10), $\hat{\mathbf{e}}$ is a point on the projected line segment, and the norm in (16) is the Euclidean (orthogonal) distance from the observed point to the corresponding line segment [11]. In both cases, (i) the prediction is computed using the event camera trajectory at the time of the event, $t_k$, and (ii) we assume the data association to be known, i.e., the correspondences between events and map primitives.[1] The likelihood (16) models only the error in the spatial domain, and not in the temporal domain since the latter is negligible: event timestamps have an accuracy in the order of a few dozen microseconds.

Following similar steps as those in (13)-(16) (independence and Gaussian error assumptions), the second and third terms in (12) lead to

$$\log\big(p(\mathcal{W}|\mathbf{x}(t))\big) = \tilde{K}_2 - \frac{1}{2}\sum_j \frac{1}{\sigma_\omega^2}\|\boldsymbol{\omega}_j - \hat{\boldsymbol{\omega}}_j(\mathbf{x}(t_j))\|^2, \tag{17}$$

$$\log\big(p(\mathcal{A}|\mathbf{x}(t))\big) = \tilde{K}_3 - \frac{1}{2}\sum_j \frac{1}{\sigma_a^2}\|\mathbf{a}_j - \hat{\mathbf{a}}_j(\mathbf{x}(t_j))\|^2, \tag{18}$$

where $\hat{\boldsymbol{\omega}}_j$ and $\hat{\mathbf{a}}_j$ are predictions of the angular velocity and linear acceleration of the event camera computed using the modeled trajectory $\mathbf{x}(t)$, such as those given by (7) and (8) in the case of a cubic spline trajectory.

Collecting terms (16)-(18), the maximization of the likelihood (12), or equivalently, its logarithm, leads to the minimization of the objective function

$$\begin{aligned}F \doteq &\frac{1}{N}\sum_{k=1}^{N}\frac{1}{\sigma_e^2}\|\mathbf{e}_k - \hat{\mathbf{e}}_k(\mathbf{x}(t_k), \mathcal{M})\|^2 \\ &+ \frac{1}{M}\sum_{j=1}^{M}\frac{1}{\sigma_\omega^2}\|\boldsymbol{\omega}_j - \hat{\boldsymbol{\omega}}_j(\mathbf{x}(t_j))\|^2 \\ &+ \frac{1}{M}\sum_{j=1}^{M}\frac{1}{\sigma_a^2}\|\mathbf{a}_j - \hat{\mathbf{a}}_j(\mathbf{x}(t_j))\|^2,\end{aligned} \tag{19}$$

where we omitted unnecessary constants. The first sum comprises the visual errors measured in the image plane and the last two sums comprise the inertial errors.

### B. Parametric Trajectory Optimization

The objective function (19) is optimized with respect to the event camera trajectory $\mathbf{x}(t)$, which in general is represented by an arbitrary curve in $SE(3)$, i.e., a "point" in an infinite-dimensional function space. However, because we represent the curve in terms of a finite set of known temporal basis functions (B-splines, formalized in (4)), the trajectory is parametrized by control poses $\mathtt{T}_{w,i}$ and, therefore, the optimization problem becomes finite dimensional. In particular, it is a non-linear least squares problem, for which standard numerical solvers such as Gauss-Newton or Levenberg-Marquardt can be applied.

In addition to the control poses, we optimize with respect to model parameters $\boldsymbol{\theta} = (\mathbf{b}_\omega^\top, \mathbf{b}_a^\top, s, \mathbf{o}^\top)^\top$, consisting of the IMU biases $\mathbf{b}_\omega$ and $\mathbf{b}_a$, and the map scale $s$ and orientation with respect to the gravity direction $\mathbf{o}$. The map orientation $\mathbf{o}$ is composed of roll and pitch angles, $\mathbf{o} = (\alpha, \beta)^\top$. Maps obtained by monocular systems, such as [12], [22], lack information about absolute map scale and orientation, so it is necessary to estimate them in such cases. We estimate the trajectory and additional model parameters by minimizing (19),

$$\{\mathtt{T}_{w,i}^*, \boldsymbol{\theta}^*\} = \arg\min_{\mathtt{T}, \boldsymbol{\theta}} F. \tag{20}$$

This optimization problem is solved in an iterative way using the Ceres solver [37], an efficient numerical implementation for non-linear least squares problems.

*Remark:* the inertial predictions are computed as described in (7) and (8), using $\mathtt{T}_{w,s}$ and its derivatives, whereas the visual predictions require the computation of $\mathtt{T}_{w,s}^{-1}$. More specifically, for each event $e_k$, triggered at time $t_k$ in the interval $[t_i, t_{i+1})$, we compute its pose $\mathtt{T}_{w,s}(u_k)$ using (4), where $u_k = (t_k - t_i)/\Delta t$. We then project the map point or line segment into the current image plane using projection matrices $\mathtt{P}(t_k) \propto \mathtt{K}(\mathtt{I}|\mathbf{0})\mathtt{T}_{w,s}^{-1}(t_k)$, $\mathtt{K}$ being the intrinsic parameter matrix of the event camera (after radial distortion compensation), and compute the distance between the event location $\mathbf{e}_k$ and the corresponding point $\hat{\mathbf{e}}_k$ in the projected primitive. To take into account the map scale $s$ and orientation $\mathbf{o}$, we right-multiply $\mathtt{P}(t_k)$ by a similarity transformation with scale $s$ and rotation $\mathtt{R}(\mathbf{o})$ before projecting the map primitives.

---

[1] In practice, we solve the data association using event-based pose-tracking algorithms that we run as a preprocessing step. Note that these algorithms rely only on the events. Details are provided in the experiments of Section VII.

## VII. Experiments

We evaluate our method on several datasets using the two different map representations in Section V: lines-based maps and point-based maps. These two representations allow us to evaluate the effect of two different visual error terms, which are the line-to-point distance and point-to-point reprojection errors presented in Section VI-A. In both cases, we quantify the trajectory accuracy using the ground truth of a motion-capture system.[2] We use the same hand-eye calibration method as described in [38]. Having a monocular setup, the absolute scale is not observable from visual observations alone. However, we are able to estimate the absolute scale since the fused IMU measurements grant scale observability. The following two sections describe the experiments with line-based maps and point-based maps, respectively. In these experiments, we used $\sigma_e = 0.1\,\text{pixel}$, $\sigma_\omega = 0.03\,\text{rad/s}$, and $\sigma_a = 0.1\,\text{m}^2/\text{s}$. We chose the values for the standard deviations of the inertial measurements to be around ten times higher than those measured at rest.

### A. Camera Trajectory Estimation in Line-based Maps

These experiments are similar to the ones presented in [11]. Here, however, we use the DAVIS instead of the DVS, which provides the following advantages. First, it has a higher spatial resolution of $240 \times 180$ pixels (instead of $128 \times 128$ pixels). Second, it provides inertial measurements (at $1\,\text{kHz}$) that are time-synchronized with the events. Third, it also outputs global-shutter intensity images (at $24\,\text{Hz}$) that we use for initialization, visualization, and a more accurate intrinsic camera calibration than that achieved using events.

*1) Tracking Method:* This method tracks a set of lines in a given metric map. Event-based line tracking is done using [19], which also provides data association between events and lines. This data association is used in the optimization of (20). Events that are not close to any line (such as the keyboard and mouse in Fig. 4a) are not considered to be part of the map and, therefore, are ignored in the optimization. Pose estimation is then done using the Gold Standard PnP algorithm [39, p.181] on the intersection points of the lines. Fig. 4 shows tracking of two different shapes.

*2) Experiment:* We moved the DAVIS sensor by hand in a motion-capture system above a square pattern, as shown in Fig. 5a. The corresponding error plots in position and orientation are shown in Fig. 5b: position error is measured using the Euclidean distance, whereas orientation error is measured using the geodesic distance in $SO(3)$ (the angle of the relative rotation between the true rotation and the estimated one) [40]. The excitation in each degree of freedom and the corresponding six error plots are shown in Fig. 9. The error statistics are summarized in Table I.

We compare three algorithms against ground truth from a motion-capture system: ($i$) the event-based tracking algorithm in [19] (in cyan), ($ii$) the proposed spline-based optimization

[2]We use a NaturalPoint OptiTrack system with 14 motion-capture cameras spanning a volume of $100\,\text{m}^3$. The system reported a calibration accuracy of $0.105\,\text{mm}$ and provides measurements at $200\,\text{Hz}$.

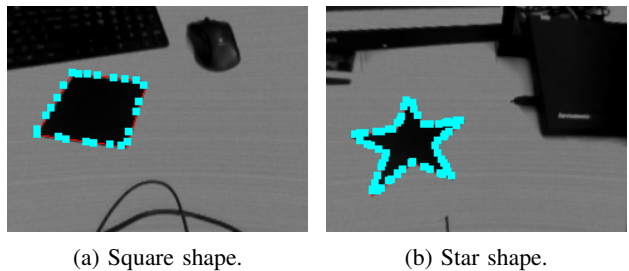

(a) Square shape.   (b) Star shape.

Fig. 4: Screenshots of the line-based tracking algorithm. The lines and the events used for its representation are in red and cyan, respectively. The image is only used for initialization and visualization.

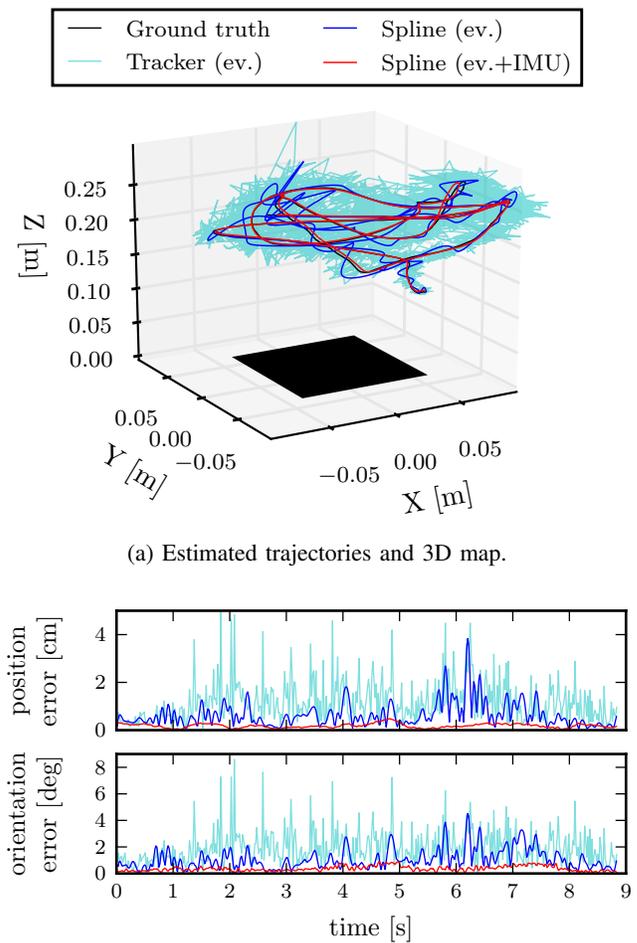

(a) Estimated trajectories and 3D map.

(b) Trajectory error in position and orientation. Legend as in Fig. 5a.

Fig. 5: Results on Line-based Tracking and Pose Estimation.

TABLE I: Results on Line-based Tracking and Pose Estimation. Position and orientation errors.

| Method | Align | Position error (abs. [cm] and rel. [%]) | | | | | | Orientation error [°] | | |
|---|---|---|---|---|---|---|---|---|---|---|
| | | $\mu$ | % | $\sigma$ | % | max | % | $\mu$ | $\sigma$ | max |
| Tracker (ev.) [19] | SE(3) | 1.11 | 3.44 | 0.75 | 2.32 | 7.96 | 24.66 | 1.87 | 1.13 | 9.61 |
| Spline (ev.) | SE(3) | 0.64 | 1.98 | 0.51 | 1.57 | 3.85 | 11.92 | 1.08 | 0.75 | 4.55 |
| Spline (ev.+IMU) | SE(3) | 0.18 | 0.57 | 0.09 | 0.27 | 0.48 | 1.48 | 0.36 | 0.19 | 0.92 |

Relative errors are given with respect to the mean scene depth.

without IMU measurements (blue color), and ($iii$) the proposed spline-based optimization (with IMU measurements, in red color). As it can be seen in the figures and in Table I, the proposed spline-based optimization ("`Spline (ev.+IMU)`" label) is more accurate than the event-based tracking algorithms: the mean, standard deviation, and maximum errors in both position and orientation are the smallest among all methods (last row of Table I). The mean position error is $0.5\,\%$ of the average scene depth and the mean orientation error is $0.37°$. The errors are up to five times smaller compared to the event-based tracking method (cf. rows 1 and 3 in Table I). Hence, the proposed method is very accurate. The benefit of including the inertial measurements in the optimization is also reported: the vision-only spline-based optimization method is better than the event-based tracking algorithm [19] (by approximately a factor of $1.5$). However, when inertial measurements are included in the optimization the errors are reduced by a factor of 4 approximately (by comparing rows 2 and 3 of Table I). Therefore, there is a significant gain in accuracy ($\times 4$ in this experiment) due to the fusion of inertial and event measurements to estimate the sensor's trajectory.

For this experiment, we placed control poses every $0.1\,\text{s}$, which led to ratios of about $5000$ events and $100$ inertial observations per control pose. We initialized the control poses by fitting a spline trajectory through the initial tracker poses.

*Scale Estimation:* In further experiments, we also estimate the absolute scale $s$ of the map as an additional parameter. As we know the map size precisely, we report the relative error. The square shape has a side length of $10\,\text{cm}$. For these experiments, we set the initial length to $0.1\,\text{cm}$, $1\,\text{cm}$, $1\,\text{m}$, and $10\,\text{m}$ (two orders of magnitude difference in both directions). The optimization converged to virtually the same minimum and the relative error was below $7\,\%$ for all cases. This error is in the same ballpark as the magnitude error of the IMU, which we measured to be about $5\,\%$ ($10.30\,\text{m/s}^2$ instead of $9.81\,\text{m/s}^2$ when the sensor is at rest).

### B. Camera Trajectory Estimation in Point-based Maps

The following experiments show that the proposed continuous-time camera trajectory estimation also works on natural scenes, i.e., without requiring strong artificial gradients to generate the events. For this, we used three sequences from the Event-Camera Dataset [38], which we refer to as *desk*, *boxes* and *dynamic* (see Figs. 6a, 7a, and 8a). The *desk* scene features a desktop with some office objects (books, a screen, a keyboard, etc.); the *boxes* scene features some boxes on a carpet, and the *dynamic* scene consists of a desk with objects and a person moving them. All datasets were recorded hand-held and contain data from the DAVIS (events, frames, and inertial measurements) as well as ground-truth pose measurements from a motion-capture system (at $200\,\text{Hz}$). We processed the data with EVO [12], an event-based visual odometry algorithm, which also returns a point-based map of the scene. Then, we used the events and the point-based map of EVO for camera trajectory optimization in the continuous-time framework, showing that we achieve higher accuracy and a smoother trajectory.

*1) Tracking Method:* EVO [12] returns both a map [41], [42] and a set of 6-DOF discrete, asynchronous poses of

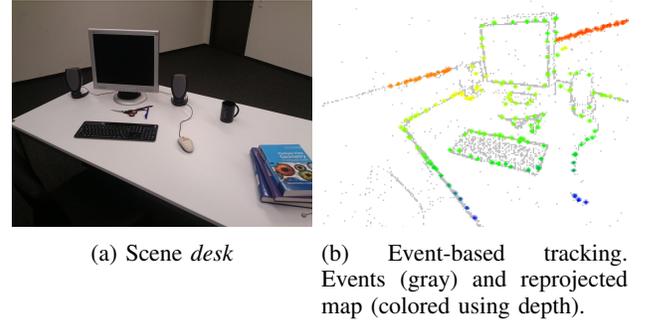

(a) Scene *desk*  (b) Event-based tracking. Events (gray) and reprojected map (colored using depth).

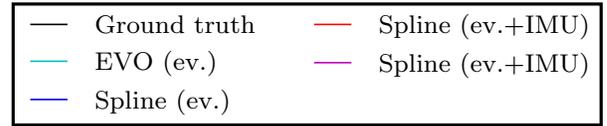

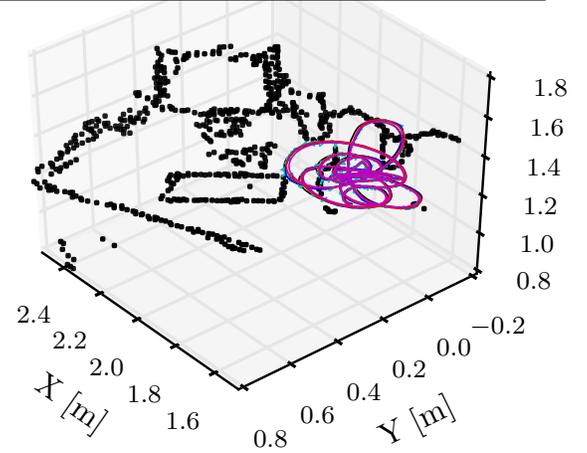

(c) Estimated trajectories and 3D map.

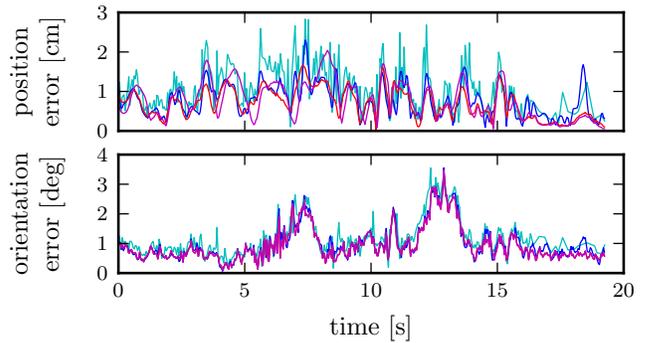

(d) Trajectory error in position and orientation. Legend as in Fig. 6c.

Fig. 6: Results for *desk* dataset.

TABLE II: Results for *desk* dataset.

| Method | Align | Position error (abs. [cm] and rel. [%]) | | | | | | Orientation error [°] | | |
|---|---|---|---|---|---|---|---|---|---|---|
| | | $\mu$ | % | $\sigma$ | % | max | % | $\mu$ | $\sigma$ | max |
| EVO (ev.) | Sim(3) | 1.08 | 0.54 | 0.53 | 0.27 | 4.64 | 2.32 | 1.31 | 0.68 | 3.55 |
| Spline (ev.) | Sim(3) | 0.78 | 0.39 | 0.40 | 0.20 | 2.30 | 1.15 | 0.98 | 0.58 | 3.56 |
| Spline (ev.+IMU) | Sim(3) | 0.69 | 0.35 | 0.36 | 0.18 | 1.66 | 0.83 | 0.94 | 0.57 | 3.47 |
| Spline (ev.+IMU) | SE(3) | 0.77 | 0.38 | 0.46 | 0.23 | 2.04 | 1.02 | 0.94 | 0.57 | 3.47 |

Relative errors are given with respect to the mean scene depth.

the event camera. In a post-processing step, we extracted the correspondences between the events and the map points that are required to optimize (20). We project the map points onto the image plane for each EVO pose and establish a correspondence if a projected point and an event are present in the same pixel. Events that cannot be associated with a map point are treated as noise and are therefore ignored in the optimization. Figs. 6b, 7b, and 8b show typical point-based maps produced by EVO, projected onto the image plane and colored according to depth with respect to the camera. The same plots also show all the observed events, colored in gray. Notice that the projected map is aligned with the observed events, as expected from an accurate tracking algorithm. The corresponding scenes are shown in Figs. 6a, 7a, and 8a.

*2) Experiments:* Figs. 6–8 and Tables II–IV summarize the results obtained on the three datasets. Figs. 6c, 7c, and 8c show the 3D maps and the event camera trajectories. Figs. 6d, 7d and 8d show the position and orientation errors obtained by comparing the estimated camera trajectories against motion-capture ground truth. Error statistics are provided in Tables II, III and IV. In additional plots in the Appendix (Figs. 9 to 12), we show the individual trajectory DOFs and their errors.

We compare four methods against ground truth from the motion-capture system: (*i*) event-based pose tracking using EVO (in cyan color in the figures), (*ii*) spline-based trajectory optimization without IMU measurements (in blue color), (*iii*) spline-based trajectory optimization (events and inertial measurements, in red color), and (*iv*) spline-based trajectory and absolute scale optimization (in magenta). The output camera trajectory of each of the first three methods was aligned with respect to ground truth using a 3D similarity transformation (rotation, translation, and uniform scaling); thus, the absolute scale is externally provided. Although a Euclidean alignment suffices (rotation and translation, without scaling) for the spline-based approach with events and IMU, we also used a similarity alignment for a fair comparison with respect to other methods. The fourth method has the same optimized camera trajectory as the third one, but the alignment with respect to the ground truth trajectory is Euclidean (6-DOF): the absolute scale is recovered from the inertial measurements. As it can be seen in Tables II, III, and IV, the spline-based approach without inertial measurements consistently achieves smaller errors than EVO (cf. rows 1 and 2 of the tables). Using also the inertial measurements further improves the results (cf. rows 2 and 3 of the tables). When using the estimated absolute map scale, the results are comparable to those where the scale was provided by ground-truth alignment with a similarity transform, even though a low-cost IMU was used (cf. rows 3 and 4 of the tables). In such a case, the mean position error is less than $1.05\,\%$ of the average scene depth, and the mean orientation error is less than $1.03°$. The standard deviations of the errors are also very small: less than $0.43\,\%$ and less than $0.57°$, respectively, in all datasets. The results are remarkably accurate. The gain in accuracy due to incorporating the inertial measurements in the optimization (with respect to the visual-only approach) is less than a factor of two, which is not as large as in the case of line-based maps (a factor of four) because EVO [12] already provides very good results compared with the line-based tracker of [19]. Nevertheless, the gain is still significant, making the event-inertial optimization consistently outperforming the event-only one.

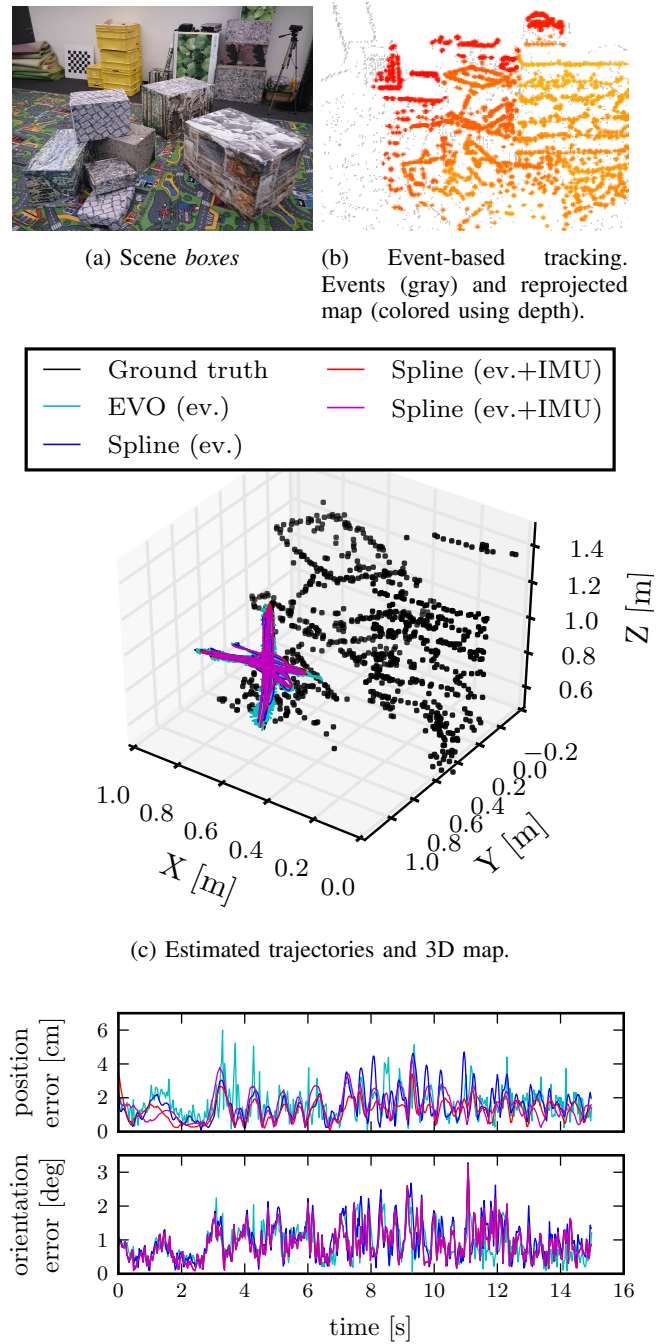

(a) Scene *boxes*  (b) Event-based tracking. Events (gray) and reprojected map (colored using depth).

(c) Estimated trajectories and 3D map.

(d) Trajectory error in position and orientation. Legend as in (c).

Fig. 7: Results for *boxes* dataset.

TABLE III: Results for *boxes* dataset.

| Method | Align | Position error (abs. [cm] and rel. [%]) | | | | | | Orientation error [°] | | |
|---|---|---|---|---|---|---|---|---|---|---|
| | | $\mu$ | % | $\sigma$ | % | max | % | $\mu$ | $\sigma$ | max |
| EVO (ev.) | Sim(3) | 1.66 | 0.61 | 0.88 | 0.33 | 6.83 | 2.51 | 0.99 | 0.50 | 2.77 |
| Spline (ev.) | Sim(3) | 1.58 | 0.58 | 0.89 | 0.33 | 4.72 | 1.73 | 0.99 | 0.54 | 3.28 |
| Spline (ev.+IMU) | Sim(3) | 1.24 | 0.46 | 0.59 | 0.22 | 3.59 | 1.32 | 0.88 | 0.48 | 3.23 |
| Spline (ev.+IMU) | SE(3) | 1.50 | 0.55 | 0.78 | 0.29 | 4.30 | 1.58 | 0.88 | 0.48 | 3.23 |

Relative errors are given with respect to the mean scene depth.

In the continuous-time methods we placed knots (the timestamps of the control poses $\mathtt{T}_{w,i}$) every $0.2\,\text{s}$, $0.15\,\text{s}$, and $0.15\,\text{s}$ for the *desk*, *boxes*, and *dynamic* datasets, respectively.

TABLE V: Dataset statistics of the Optimization (20)

| Experiment | # Events | # IMU | # Control Poses | Duration [s] |
|---|---|---|---|---|
| line-based | 450,416 | 8,842 | 92 | 8.8 |
| desk | 883,449 | 19,317 | 99 | 19.3 |
| boxes | 2,064,028 | 14,977 | 103 | 15.0 |
| dynamic | 879,143 | 14,976 | 103 | 15.0 |

TABLE VI: Computational cost of the Optimization (20)

| | Events-only | | Events + IMU | |
|---|---|---|---|---|
| | Time [s] | Iterations | Time [s] | Iterations |
| line-based | 28.4 | 2 | 48.0 | 3 |
| desk | 110.7 | 8 | 91.4 | 3 |
| boxes | 82.0 | 1 | 182.2 | 3 |
| dynamic | 34.7 | 1 | 65.4 | 2 |

This led to ratios of about $10^4$ events and 150–200 inertial measurements per control pose. We initialized the control poses by fitting a spline through the initial tracker poses.

*3) Absolute Map Scale and Gravity Alignment:* In the above experiments with IMU, we also estimated the absolute scale $s$ and orientation $\mathbf{o}$ of the map as additional parameters. Since EVO is monocular, it cannot estimate the absolute scale. However, by fusing the inertial data with EVO, it is possible to recover the absolute scale and to align the map with gravity. We found that the absolute scale deviated from the true value by $4.1\,\%$, $6.5\,\%$, and $2.8\,\%$ for the *desk*, *boxes*, and *dynamic* datasets, respectively. For the alignment with gravity, we found that the estimated gravity direction deviated from the true value by $3.83°$, $20.18°$, and $3.34°$ for the *desk*, *boxes*, and *dynamic* datasets, respectively. The high alignment error for the *boxes* dataset is likely due to the dominant translational motion of the camera (i.e., lack of a rich rotational motion).

### C. Computational Cost

Table VI reports the runtime for the least-squares optimization of (20) using the Ceres library [37] and the number of iterations taken to converge to a tolerance of $10^{-3}$ in the change of the objective function value. Table V provides an overview of the experiments (dataset duration, number of events and inertial measurements, and number of control poses used). The experiments were conducted on a laptop with an Intel Core i7-3720QM CPU at $2.60\,\text{GHz}$.

The optimization process typically converges within a few iterations (ten or less). Depending on the number of iterations, our approach is around 3 to 13 times slower than real-time. Most of the computation time (around $80\,\%$) is devoted to the evaluation of Jacobians, which is done using automatic differentiation. The optimization could be made real time by adding more computational power (such as a GPU), by following a sliding-window approach (i.e., using only the most recent history of measurements), or by using analytical derivatives and approximations, such as using the same pose derivative for several measurements that are close in time.

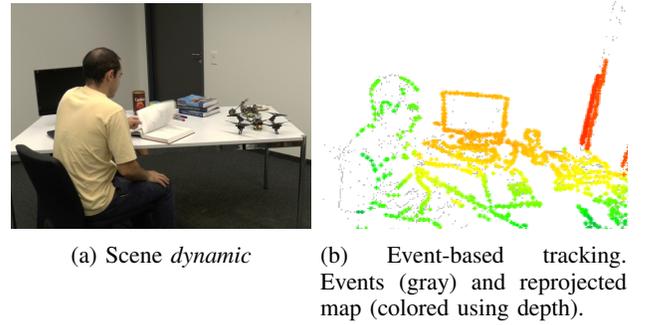

(a) Scene *dynamic*  (b) Event-based tracking. Events (gray) and reprojected map (colored using depth).

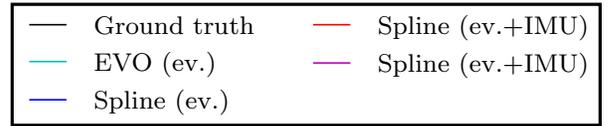

— Ground truth — Spline (ev.+IMU)
— EVO (ev.) — Spline (ev.+IMU)
— Spline (ev.)

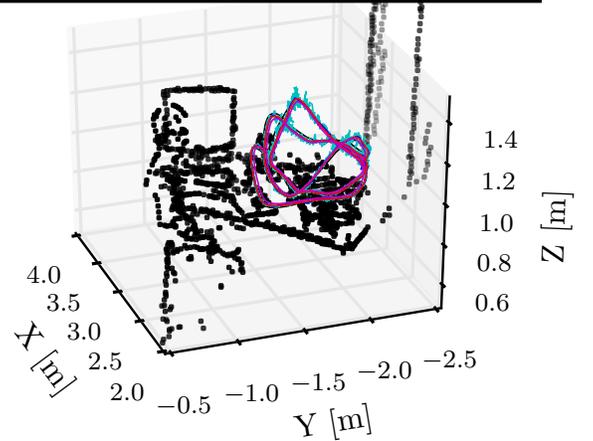

(c) Estimated trajectories and 3D map.

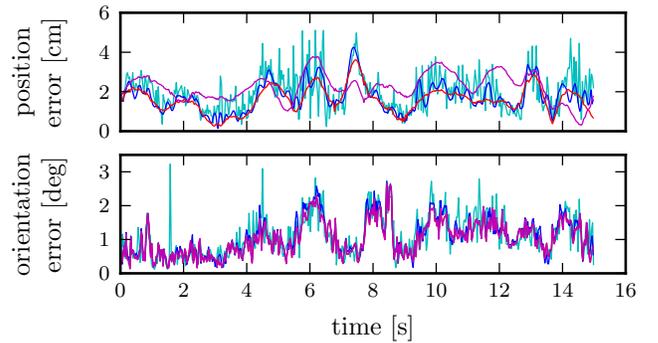

(d) Trajectory error in position and orientation. Legend as in (c).

Fig. 8: Results for *dynamic* dataset.

TABLE IV: Results for *dynamic* dataset.

| Method | Align | Position error (abs. [cm] and rel. [%]) | | | | | | Orientation error [°] | | |
|---|---|---|---|---|---|---|---|---|---|---|
| | | $\mu$ | % | $\sigma$ | % | max | % | $\mu$ | $\sigma$ | max |
| EVO (ev.) | Sim(3) | 1.94 | 1.43 | 0.94 | 0.69 | 7.06 | 5.20 | 1.08 | 0.58 | 3.66 |
| Spline (ev.) | Sim(3) | 1.74 | 1.28 | 0.74 | 0.54 | 6.40 | 4.71 | 1.08 | 0.56 | 3.42 |
| Spline (ev.+IMU) | Sim(3) | 1.60 | 1.17 | 0.65 | 0.47 | 3.65 | 2.68 | 1.02 | 0.51 | 3.43 |
| Spline (ev.+IMU) | SE(3) | 2.22 | 1.63 | 0.69 | 0.51 | 3.79 | 2.79 | 1.02 | 0.51 | 3.43 |

Relative errors are given with respect to the mean scene depth.

## VIII. Discussion

Event cameras provide visual measurements asynchronously and at very high rate. Traditional formulations, which describe the camera trajectory using poses at discrete timestamps, are not appropriate to deal with such almost continuous data streams because of the difficulty in establishing correspondences between the discrete sets of events and poses, and because the preservation of the temporal information of the events would require a very large number of poses (one per event). The continuous-time framework is a convenient representation of the camera trajectory since it has many desirable properties, among them: (i) it solves the issue of establishing correspondences between events and poses (since the pose at the time of the event is well-defined), and (ii) it is a natural framework for data fusion: it deals with the asynchronous nature of the events as well the synchronous samples from the IMU. As demonstrated in the experiments, such event-inertial data fusion allows significantly increasing the accuracy of the estimated camera motion over event-only–based approaches (e.g., by a factor of four).

The proposed parametric B-spline model makes the trajectory optimization computationally feasible since it has local basis functions (i.e., sparse Hessian matrix), analytical derivatives (i.e., fast to compute), and it is a compact representation: few parameters (control poses) suffice to assimilate several hundred thousand events and inertial measurements while providing a smooth trajectory. Our method demonstrated its usefulness to refine trajectories from state-of-the-art event-based pose trackers such as EVO, with or without inertial measurements. The current implementation of our method runs off-line, as a full smoothing post-processing stage (taking into account the full history of measurements), but the method can be adapted for on-line processing in a fixed-lag smoothing manner (i.e., using only the most recent history of measurements); the local support of the B-spline basis functions enables such type of local temporal processing. However, due to the high computational requirements of the current implementation, further optimizations and approximations need to be investigated to achieve real-time performance.

Another reason for adopting the continuous-time trajectory framework is that it is agnostic to the map representation. We showed that the proposed method is flexible, capable of estimating accurate camera trajectories in scenes with line-based maps as well as point-based maps. In fact, the probabilistic (maximum likelihood) justification of the optimization approach gracefully unifies both formulations, lines and points, under the same objective function in a principled way. In this manner, we extended the method in [11] and broadened its applicability to different types of maps (i.e., scenes). The probabilistic formulation also allows a straightforward generalization to other error distributions besides the normal one. More specifically, the results of the proposed method on line-based and point-based maps show similar remarkable accuracy, with mean position error of less than $1\,\%$ of the average scene depth, and mean orientation error of less than $1°$. The absolute scale and gravity direction are recovered in both types of maps, with an accuracy of approximately $5\,\%$, which matches the accuracy of the IMU accelerometer; thus, the proposed method takes full advantage of the accuracy of the available sensors.

Using cumulative B-splines in $SE(3)$ sets a prior on the shape of the camera trajectory. While smooth rigid-body motions are well-approximated with such basis functions, they are not suitable to fit discontinuities (such as bumps or crashes). In this work, we use fixed temporal spacing of the control poses, which is not optimal when the motion speed changes abruptly within a dataset. Choosing the optimal number of control poses and their temporal spacing is beyond the scope of this paper and is left for future work.

## IX. Conclusion

In this paper, we presented a visual-inertial odometry method for event cameras using a continuous-time framework. This approach can deal with the asynchronous nature of the events and the high frequency of the inertial measurements in a principled way while providing a compact and smooth representation of the trajectory using a parametric model. The sensor's trajectory is approximated by a smooth curve in the space of rigid-body motions using cubic splines. The approximated trajectory is then optimized according to a geometrically meaningful error measure in the image plane and a direct inclusion of the inertial measurements, which have probabilistic justifications. We tested our method on real data from two recent algorithms: a simple line-based tracker and an event-based visual odometry algorithm that works on natural scenes. In all experiments, our method outperformed previous algorithms when comparing to ground truth, with a remarkable accuracy: mean position error of less than $1\,\%$ of the average scene depth, and mean orientation error of less than $1°$.

## Appendix A
## Analytical Derivatives of the Camera Trajectory

By the chain rule, the first and second temporal derivatives of the spline trajectory (4) are, using Newton's dot notation for differentiation,

$$\dot{\mathtt{T}}_{w,s}(u) = \mathtt{T}_{w,i-1} \begin{pmatrix} \dot{\mathbf{A}}_0 \mathbf{A}_1 \mathbf{A}_2 \\ + \mathbf{A}_0 \dot{\mathbf{A}}_1 \mathbf{A}_2 \\ + \mathbf{A}_0 \mathbf{A}_1 \dot{\mathbf{A}}_2 \end{pmatrix}, \quad (21)$$

$$\ddot{\mathtt{T}}_{w,s}(u) = \mathtt{T}_{w,i-1} \begin{pmatrix} \ddot{\mathbf{A}}_0 \mathbf{A}_1 \mathbf{A}_2 + \mathbf{A}_0 \ddot{\mathbf{A}}_1 \mathbf{A}_2 \\ + \mathbf{A}_0 \mathbf{A}_1 \ddot{\mathbf{A}}_2 + 2\dot{\mathbf{A}}_0 \dot{\mathbf{A}}_1 \mathbf{A}_2 \\ + 2\dot{\mathbf{A}}_0 \mathbf{A}_1 \dot{\mathbf{A}}_2 + 2\mathbf{A}_0 \dot{\mathbf{A}}_1 \dot{\mathbf{A}}_2 \end{pmatrix}, \quad (22)$$

respectively, where

$$\mathbf{A}_j \doteq \exp\bigl(\Omega_{i+j} \tilde{\mathbf{B}}(u)_{j+1}\bigr), \quad (23)$$

$$\dot{\mathbf{A}}_j = \mathbf{A}_j \Omega_{i+j} \dot{\tilde{\mathbf{B}}}(u)_{j+1}, \quad (24)$$

$$\ddot{\mathbf{A}}_j = \dot{\mathbf{A}}_j \Omega_{i+j} \dot{\tilde{\mathbf{B}}}(u)_{j+1} + \mathbf{A}_j \Omega_{i+j} \ddot{\tilde{\mathbf{B}}}(u)_{j+1}, \quad (25)$$

$$\dot{\tilde{\mathbf{B}}} = \frac{1}{\Delta t} \mathbf{C} \begin{bmatrix} 0 \\ 1 \\ 2u \\ 3u^2 \end{bmatrix}, \quad \ddot{\tilde{\mathbf{B}}} = \frac{1}{\Delta t^2} \mathbf{C} \begin{bmatrix} 0 \\ 0 \\ 2 \\ 6u \end{bmatrix}. \quad (26)$$

Note that (4), (7) and (23)-(25) correct some symbols and typos in the indices of the formulas provided by [5].

## References

[1] P. Lichtsteiner, C. Posch, and T. Delbruck, "A 128×128 120 dB 15 µs latency asynchronous temporal contrast vision sensor," *IEEE J. Solid-State Circuits*, vol. 43, no. 2, pp. 566–576, 2008.

[2] C. Brandli, R. Berner, M. Yang, S.-C. Liu, and T. Delbruck, "A 240x180 130dB 3us latency global shutter spatiotemporal vision sensor," *IEEE J. Solid-State Circuits*, vol. 49, no. 10, pp. 2333–2341, 2014.

[3] C. Posch, D. Matolin, and R. Wohlgenannt, "A QVGA 143 dB dynamic range frame-free PWM image sensor with lossless pixel-level video compression and time-domain CDS," *IEEE J. Solid-State Circuits*, vol. 46, no. 1, pp. 259–275, Jan. 2011.

[4] L. Kneip, D. Scaramuzza, and R. Siegwart, "A novel parametrization of the perspective-three-point problem for a direct computation of absolute camera position and orientation," in *Proc. IEEE Int. Conf. Comput. Vis. Pattern Recog.*, 2011, pp. 2969–2976.

[5] A. Patron-Perez, S. Lovegrove, and G. Sibley, "A spline-based trajectory representation for sensor fusion and rolling shutter cameras," *Int. J. Comput. Vis.*, vol. 113, no. 3, pp. 208–219, 2015.

[6] C. Bibby and I. D. Reid, "A hybrid SLAM representation for dynamic marine environments," in *IEEE Int. Conf. Robot. Autom. (ICRA)*, 2010, pp. 257–264.

[7] P. Furgale, T. D. Barfoot, and G. Sibley, "Continuous-time batch estimation using temporal basis functions," in *IEEE Int. Conf. Robot. Autom. (ICRA)*, 2012, pp. 2088–2095.

[8] S. Anderson, F. Dellaert, and T. D. Barfoot, "A hierarchical wavelet decomposition for continuous-time SLAM," in *IEEE Int. Conf. Robot. Autom. (ICRA)*, 2014, pp. 373–380.

[9] H. S. Alismail, L. D. Baker, and B. Browning, "Continuous trajectory estimation for 3D SLAM from actuated lidar," in *IEEE Int. Conf. Robot. Autom. (ICRA)*, 2014, pp. 6096–6101.

[10] C. Kerl, J. Stückler, and D. Cremers, "Dense continuous-time tracking and mapping with rolling shutter RGB-D cameras," in *Int. Conf. Comput. Vis. (ICCV)*, 2015, pp. 2264–2272.

[11] E. Mueggler, G. Gallego, and D. Scaramuzza, "Continuous-time trajectory estimation for event-based vision sensors," in *Robotics: Science and Systems (RSS)*, 2015.

[12] H. Rebecq, T. Horstschäfer, G. Gallego, and D. Scaramuzza, "EVO: A geometric approach to event-based 6-DOF parallel tracking and mapping in real-time," *IEEE Robot. Autom. Lett.*, vol. 2, pp. 593–600, 2017.

[13] T. Delbruck, V. Villanueva, and L. Longinotti, "Integration of dynamic vision sensor with inertial measurement unit for electronically stabilized event-based vision," in *IEEE Int. Symp. Circuits Syst. (ISCAS)*, Jun. 2014, pp. 2636–2639.

[14] D. Weikersdorfer and J. Conradt, "Event-based particle filtering for robot self-localization," in *IEEE Int. Conf. Robot. Biomimetics (ROBIO)*, 2012, pp. 866–870.

[15] D. Weikersdorfer, R. Hoffmann, and J. Conradt, "Simultaneous localization and mapping for event-based vision systems," in *Int. Conf. Comput. Vis. Syst. (ICVS)*, 2013, pp. 133–142.

[16] A. Censi and D. Scaramuzza, "Low-latency event-based visual odometry," in *IEEE Int. Conf. Robot. Autom. (ICRA)*, 2014, pp. 703–710.

[17] D. Weikersdorfer, D. B. Adrian, D. Cremers, and J. Conradt, "Event-based 3D SLAM with a depth-augmented dynamic vision sensor," in *IEEE Int. Conf. Robot. Autom. (ICRA)*, Jun. 2014, pp. 359–364.

[18] B. Kueng, E. Mueggler, G. Gallego, and D. Scaramuzza, "Low-latency visual odometry using event-based feature tracks," in *IEEE/RSJ Int. Conf. Intell. Robot. Syst. (IROS)*, Daejeon, Korea, Oct. 2016, pp. 16–23.

[19] E. Mueggler, B. Huber, and D. Scaramuzza, "Event-based, 6-DOF pose tracking for high-speed maneuvers," in *IEEE/RSJ Int. Conf. Intell. Robot. Syst. (IROS)*, 2014, pp. 2761–2768.

[20] G. Gallego, J. E. A. Lund, E. Mueggler, H. Rebecq, T. Delbruck, and D. Scaramuzza, "Event-based, 6-DOF camera tracking from photometric depth maps," *IEEE Trans. Pattern Anal. Machine Intell.*, 2017.

[21] H. Kim, A. Handa, R. Benosman, S.-H. Ieng, and A. J. Davison, "Simultaneous mosaicing and tracking with an event camera," in *British Machine Vis. Conf. (BMVC)*, 2014.

[22] H. Kim, S. Leutenegger, and A. J. Davison, "Real-time 3D reconstruction and 6-DoF tracking with an event camera," in *Eur. Conf. Comput. Vis. (ECCV)*, 2016, pp. 349–364.

[23] G. Gallego and D. Scaramuzza, "Accurate angular velocity estimation with an event camera," *IEEE Robot. Autom. Lett.*, vol. 2, pp. 632–639, 2017.

[24] G. Gallego, H. Rebecq, and D. Scaramuzza, "A unifying contrast maximization framework for event cameras, with applications to motion, depth, and optical flow estimation," in *Proc. IEEE Int. Conf. Comput. Vis. Pattern Recog.*, 2018.

[25] A. I. Maqueda, A. Loquercio, G. Gallego, N. García, and D. Scaramuzza, "Event-based vision meets deep learning on steering prediction for self-driving cars," in *Proc. IEEE Int. Conf. Comput. Vis. Pattern Recog.*, 2018.

[26] A. Z. Zhu, N. Atanasov, and K. Daniilidis, "Event-based visual inertial odometry," in *Proc. IEEE Int. Conf. Comput. Vis. Pattern Recog.*, 2017, pp. 5816–5824.

[27] H. Rebecq, T. Horstschäfer, and D. Scaramuzza, "Real-time visual-inertial odometry for event cameras using keyframe-based nonlinear optimization," in *British Machine Vis. Conf. (BMVC)*, Sep. 2017.

[28] A. Rosinol Vidal, H. Rebecq, T. Horstschaefer, and D. Scaramuzza, "Ultimate SLAM? combining events, images, and IMU for robust visual SLAM in HDR and high speed scenarios," *IEEE Robot. Autom. Lett.*, vol. 3, no. 2, pp. 994–1001, Apr. 2018.

[29] C. Harris and M. Stephens, "A combined corner and edge detector," in *Proceedings of The Fourth Alvey Vision Conference*, vol. 15, 1988, pp. 147–151.

[30] E. Rosten and T. Drummond, "Machine learning for high-speed corner detection," in *Eur. Conf. Comput. Vis. (ECCV)*, 2006, pp. 430–443.

[31] E. Mueggler, C. Bartolozzi, and D. Scaramuzza, "Fast event-based corner detection," in *British Machine Vis. Conf. (BMVC)*, 2017.

[32] A. I. Mourikis and S. I. Roumeliotis, "A multi-state constraint Kalman filter for vision-aided inertial navigation," in *IEEE Int. Conf. Robot. Autom. (ICRA)*, Apr. 2007, pp. 3565–3572.

[33] S. Leutenegger, S. Lynen, M. Bosse, R. Siegwart, and P. Furgale, "Keyframe-based visual–inertial odometry using nonlinear optimization," *Int. J. Robot. Research*, vol. 34, no. 3, pp. 314–334, 2015.

[34] C. Forster, L. Carlone, F. Dellaert, and D. Scaramuzza, "On-manifold preintegration for real-time visual-inertial odometry," *IEEE Trans. Robot.*, vol. 33, no. 1, pp. 1–21, Feb 2017.

[35] Y. Ma, S. Soatto, J. Košecká, and S. S. Sastry, *An Invitation to 3-D Vision: From Images to Geometric Models*. Springer, 2004.

[36] G. Gallego and A. Yezzi, "A compact formula for the derivative of a 3-D rotation in exponential coordinates," *J. Math. Imaging Vis.*, vol. 51, no. 3, pp. 378–384, Mar. 2015.

[37] A. Agarwal, K. Mierle, and Others, "Ceres solver," ceres-solver.org.

[38] E. Mueggler, H. Rebecq, G. Gallego, T. Delbruck, and D. Scaramuzza, "The event-camera dataset and simulator: Event-based data for pose estimation, visual odometry, and SLAM," *Int. J. Robot. Research*, vol. 36, pp. 142–149, 2017.

[39] R. Hartley and A. Zisserman, *Multiple View Geometry in Computer Vision*. Cambridge University Press, 2003, second Edition.

[40] D. Q. Huynh, "Metrics for 3D rotations: Comparison and analysis," *J. Math. Imaging Vis.*, vol. 35, no. 2, pp. 155–164, 2009.

[41] H. Rebecq, G. Gallego, and D. Scaramuzza, "EMVS: Event-based multi-view stereo," in *British Machine Vis. Conf. (BMVC)*, Sep. 2016.

[42] H. Rebecq, G. Gallego, E. Mueggler, and D. Scaramuzza, "EMVS: Event-based multi-view stereo—3D reconstruction with an event camera in real-time," *Int. J. Comput. Vis.*, pp. 1–21, Nov. 2017.

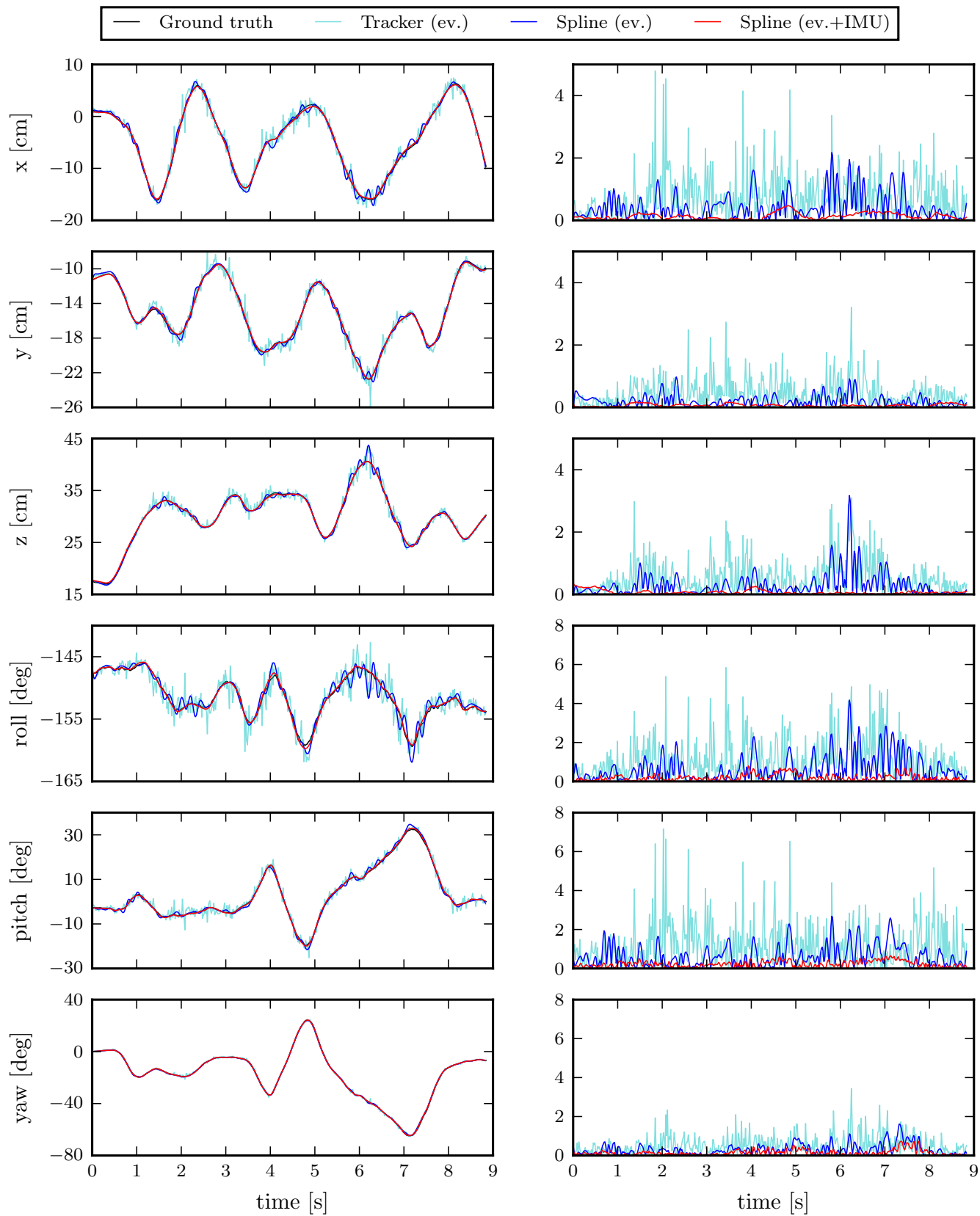

Fig. 9: *line-based* dataset. Plots of the 6-DOF (left column) and error (right column) of the estimated trajectories in Fig. 5a.

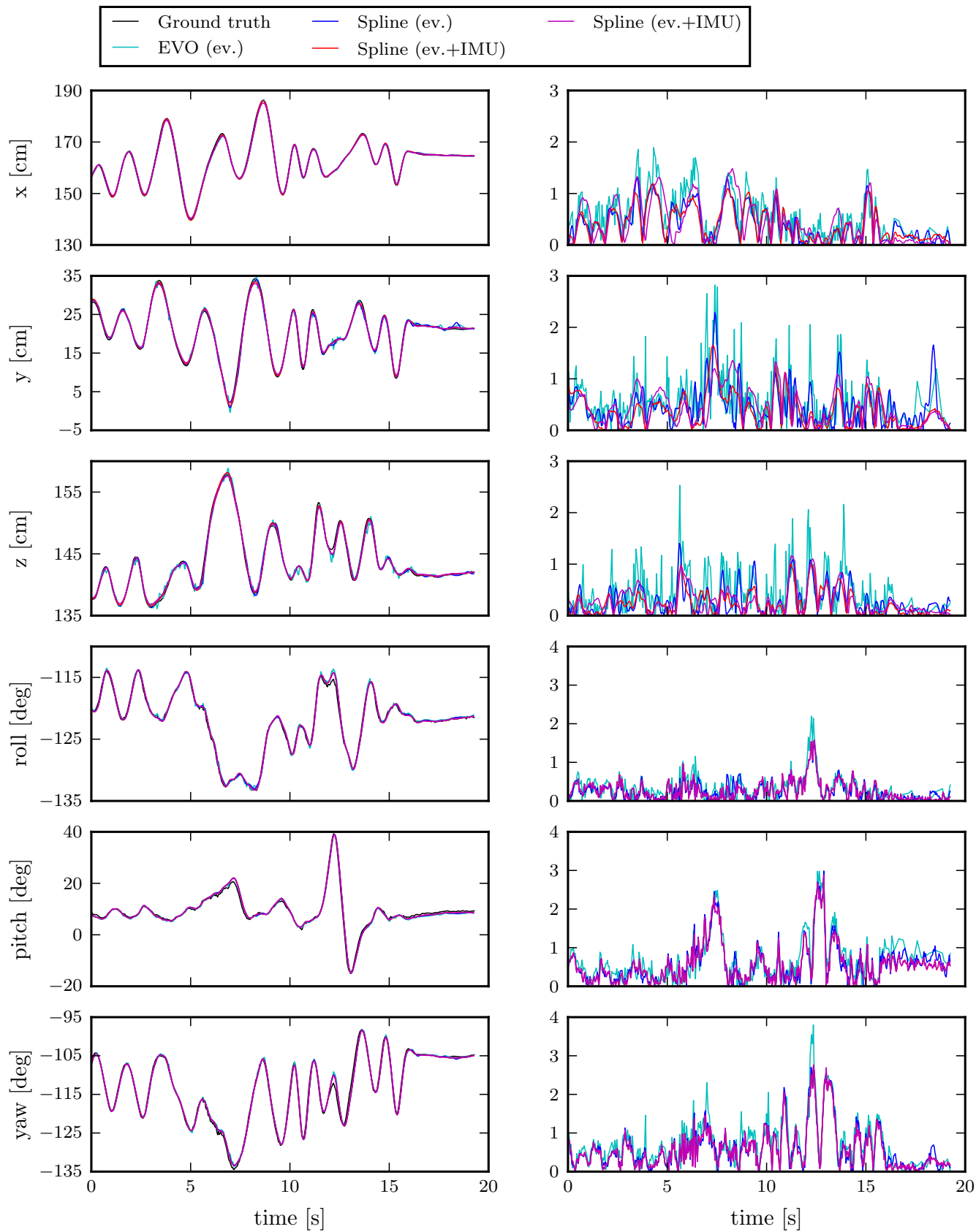

Fig. 10: *desk* dataset. Plots of the 6-DOF (left column) and error (right column) of the estimated trajectories in Fig. 6c.

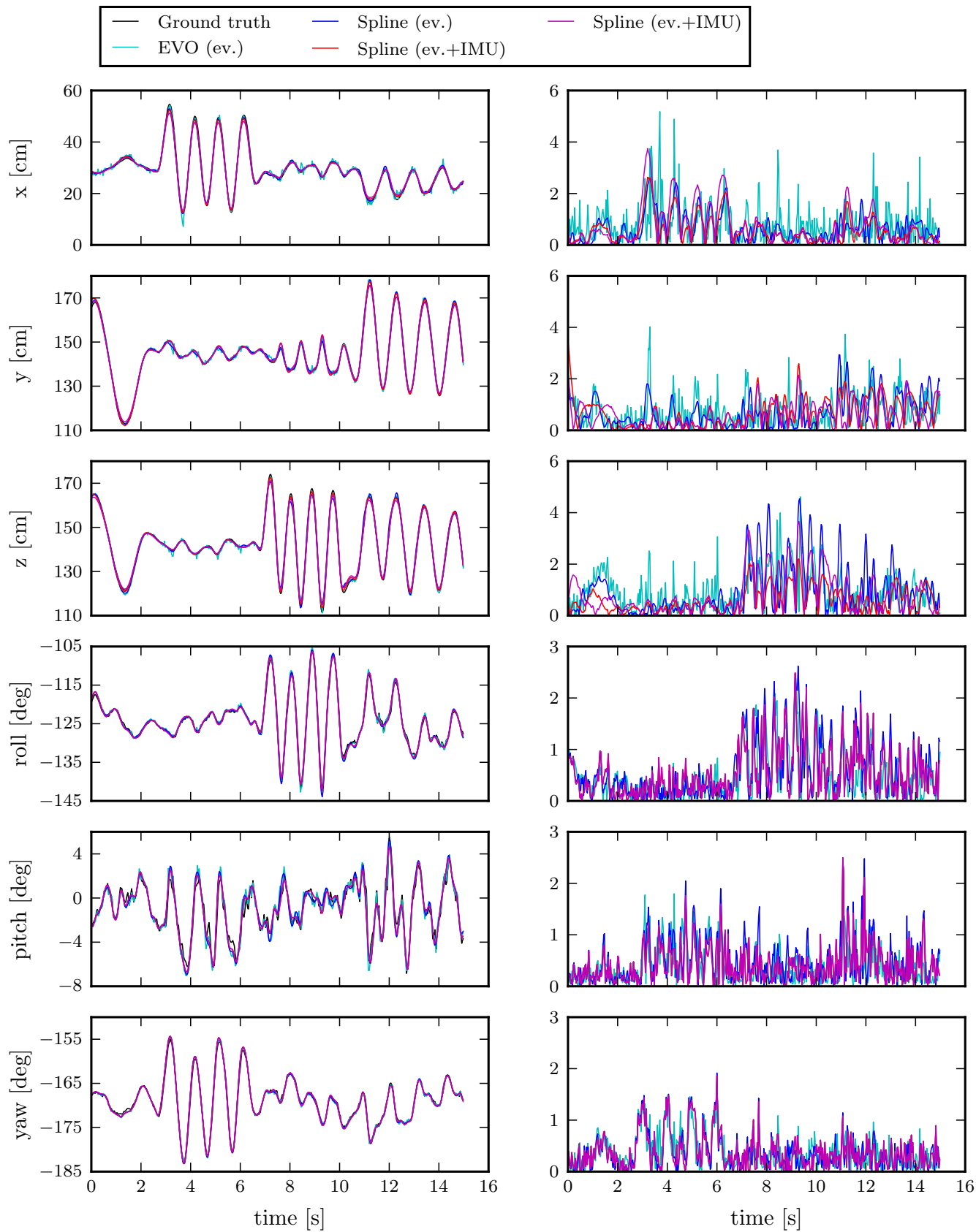

Fig. 11: *boxes* dataset. Plots of the 6-DOF (left column) and error (right column) of the estimated trajectories in Fig. 7c.

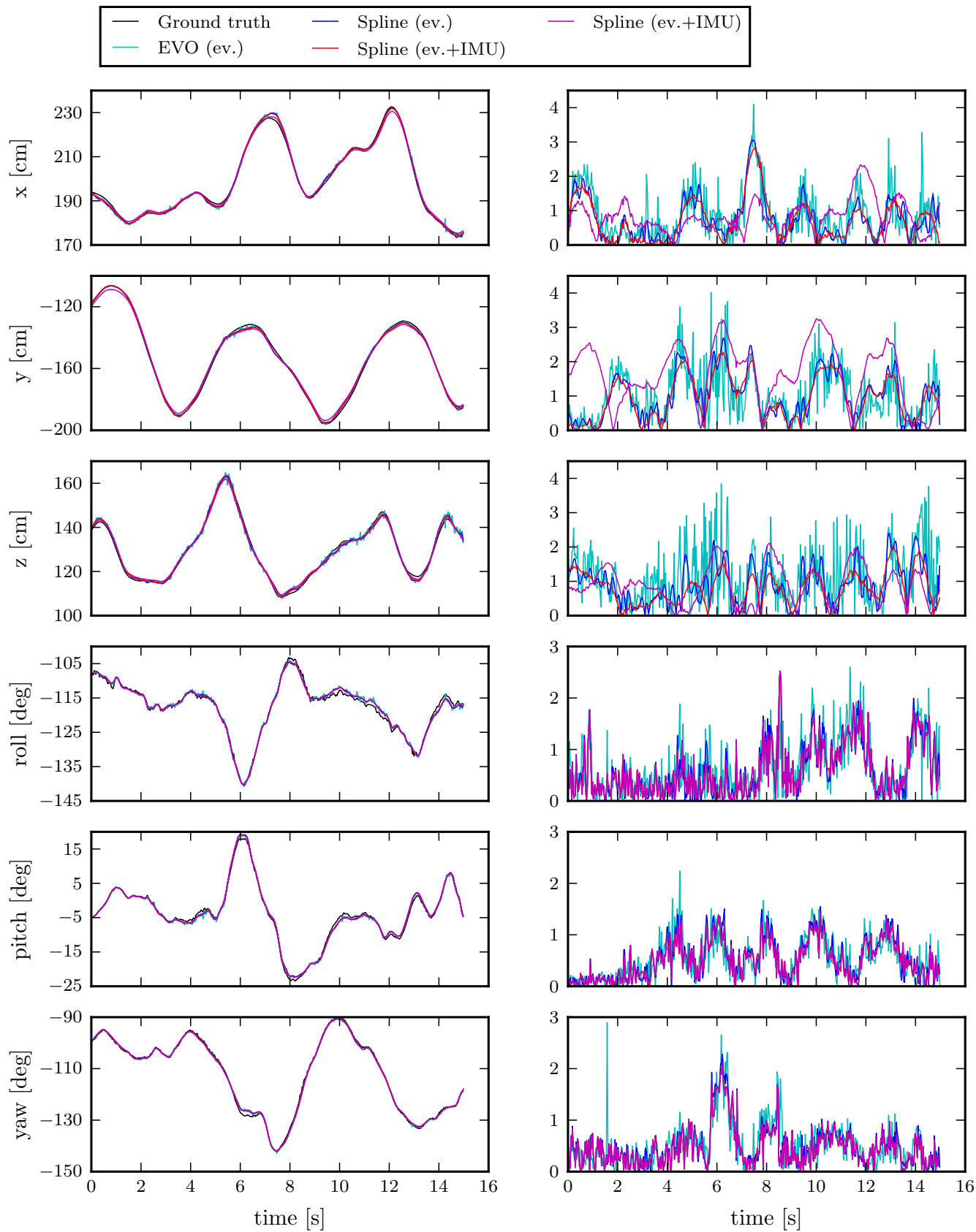

Fig. 12: *dynamic* dataset. Plots of the 6-DOF (left column) and error (right column) of the estimated trajectories in Fig. 8c